\documentclass{article}

\usepackage{PRIMEarxiv}

\usepackage[utf8]{inputenc} 
\usepackage[T1]{fontenc}    
\usepackage{hyperref}       
\usepackage{url}            
\usepackage{booktabs}       
\usepackage{amsfonts}       
\usepackage{nicefrac}       
\usepackage{microtype}      
\usepackage{lipsum}
\usepackage{fancyhdr}       
\usepackage{graphicx}       
\graphicspath{{media/}}     

\usepackage{color,soul}

\usepackage{amsmath}
\usepackage{graphicx}
\usepackage{algorithm}
\usepackage{algpseudocode}

\usepackage{amsmath}
\usepackage{graphicx}
\usepackage{algorithm}
\usepackage{algpseudocode}
\usepackage{multirow}
\usepackage{xcolor}

\algdef{SE}[DOWHILE]{Do}{doWhile}{\algorithmicdo}[1]{\algorithmicwhile\ #1}%
\algnewcommand\algorithmicinput{\textbf{Input:}}
\algnewcommand\algorithmicoutput{\textbf{Output:}}
\algnewcommand\Input{\item[\algorithmicinput]}
\algnewcommand\Output{\item[\algorithmicoutput]}

\title{Imperceptible Adversarial Attack on Deep Neural Networks from Image Boundary
}

\author{
  Fahad Alrasheedi, Xin Zhong \\
  Department of Computer Science \\
  University of Nebraska Omaha \\
  Omaha, Nebraska, USA\\
  \texttt{\{falrasheedi, xzhong\}@unomaha.edu} \\
}

\begin{document}
\maketitle

\pagestyle{empty}

\begin{abstract}
Although Deep Neural Networks (DNNs), such as the convolutional neural networks (CNN) and Vision Transformers (ViTs), have been successfully applied in the field of computer vision, they are demonstrated to be vulnerable to well-sought Adversarial Examples (AEs) that can easily fool the DNNs. 
The research in AEs has been active, and many adversarial attacks and explanations have been proposed since they were discovered in 2014. The mystery of the AE's existence is still an open question, and many studies suggest that DNN training algorithms have blind spots.
   
The salient objects usually do not overlap with boundaries; hence, the boundaries are not the DNN model's attention. 
Nevertheless, recent studies show that the boundaries can dominate the behavior of the DNN models. 
Hence, this study aims to look at the AEs from a different perspective and proposes an imperceptible adversarial attack that systemically attacks the input image boundary for finding the AEs. 
The experimental results have shown that the proposed boundary attacking method effectively attacks six CNN models and the ViT using only 32\% of the input image content (from the boundaries) with an average success rate (SR) of 95.2\% and an average peak signal-to-noise ratio of 41.37 dB. 
Correlation analyses are conducted, including the relation between the adversarial boundary's width and the SR and how the adversarial boundary changes the DNN model's attention. 
This paper's discoveries can potentially advance the understanding of AEs and provide a different perspective on how AEs can be constructed. 
\end{abstract}


\section{Introduction}
\label{submission}

Deep Neural Networks (DNNs) have successfully advanced many applications of computer vision such as image classification~\cite{resnet, vgg}, object detection~\cite{yolo,rccn,od1,od2}, saliency detection~\cite{Basnet,u2net,sod}, facial recognition~\cite{pcanet,deepid3,fr1}, and many others. 
However, such DNNs are shown to be vulnerable to adversarial attacks that can convert Clean Examples, correctly classified by a DNN model, to Adversarial Examples~\cite{L-BFGS}. 
The conversion process is a mathematical process that tweaks the Clean Example in well-chosen directions in the feature space with some energy, usually called \textit{epsilon}, until the DNN model produces a wrong output, and the total of the tweaks is called adversarial perturbation~\cite{FGSM,deepfool}. 
The adversarial perturbations can be categorized into two categories (explained in Section~\ref{sec:literature}): imperceptible to the human eye and perceptible; this study focuses on the first category.


To deceive the human vision, the invisibility of the adversarial perturbations is one of the fundamental considerations for the imperceptible adversarial perturbations. 
Hence, humans should barely notice the difference between a Clean Example and its Adversarial Example. For instance, the imperceptibility of the adversarial perturbations produced by the Fast Gradient Sign Methods~\cite{FGSM} depends on the value of \textit{epsilon} where higher values of epsilon lead to conspicuous differences between a Clean Example and its Adversarial Example. 
Hence, much research addressed the imperceptibility by introducing algorithms that iteratively increased the adversarial perturbation constrained on flipping the output of the DNN model~\cite{IFGSM,deepfool}. 
Such studies changed the whole features in the input space; however, some features were seen as needless to contribute to the conversion of the DNN model’s output. 

\begin{figure}[t!]
\includegraphics[width=8.5cm]{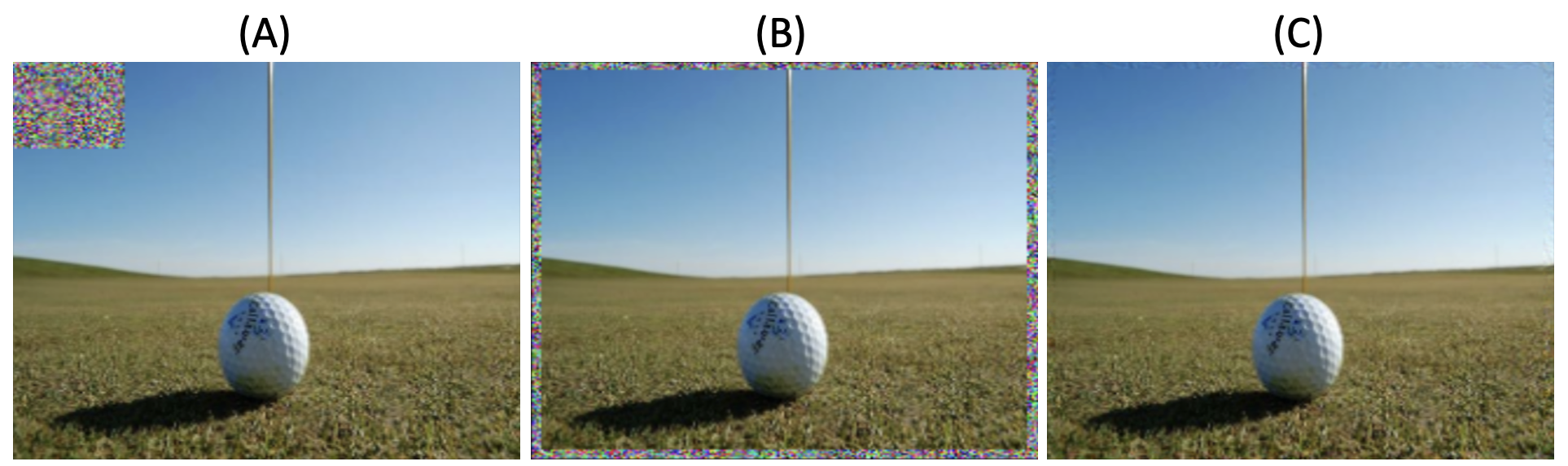} 
   \centering
   \caption{Three Adversarial Examples crafted from the same Clean Example by three different adversarial attacks: A) Patch of \cite{lavan}, B) 5-pixel Frame of \cite{edgeframe}, and C) 5-pixel Boundary of our attack. The true label is golf ball while the three attacks agree on the Adversarial Label (parachute).}
   \vspace{-1.5em}
   \label{fig:intro}
\end{figure}

Consequently, a line of research went to investigate the possibility of finding Adversarial Examples by attacking only the important features in the input space. 
Hence, various adversarial attacks were proposed in which a saliency map was employed to find and attack the salient features in the input space, such as Jacobian-based Saliency Map Attack (JSMA)~\cite{jsma}, Maximal Jacobian-based Saliency map attack (MJSMT)~\cite{mjsmt}, and Probabilistic Jacobian based Saliency map attack (PJSMT)~\cite{pjsmt}. 
Moreover, Qian \textit{et al.}~\cite{visually} proposed an adversarial attack that used an attention model to find a small continuous area, called a contributing feature region (CFR), in the input to attack; they called that perturbation \textit{imperceptible adversarial patch}. 
Such studies produced imperceptible perturbations; however, their arguments for successfully finding the Adversarial Examples were based on finding and attacking important features in the input space and ignoring the other features that were seen as unimportant in the adversarial attacks.

Unimportant features could still be critical to cause Adversarial Examples. Hence, it is possible to only modify a small area, and include unimportant features in the input space to change a Clean Example to an Adversarial Example. 
Here, the modifications usually came as adversarial patches~\cite{adversarailPatch,lavan,Advhat} or adversarial frames~\cite{edgeframe, edgeframe} that covered an unimportant area and were able to make that area contributing to find the Adversarial Examples. Also, such patches or frames could partially cover the important features in the input space and make the DNN's model classify the input adversarially~\cite{glassframe,adversarailPatch}.  
Moreover, Su \textit{et al.}~\cite{onepixel}introduced the application of Differential Evolution (DE) to find the Adversarial Examples by changing several features ranging from one to five features. 
Usually, the values of those DE-selected features extremely contrast with the neighboring features; the DE algorithm usually, selected from the important features; but it also could choose features that were not overlapped with the main object in the input. These studies usually lead to salient noises that can be noticed by human vision easily.

This study is motivated to investigate the possibility of attacking the DNN's models using only unimportant features in the input space while still keeping the adversarial perturbation imperceptible to the human vision as shown in Figure\ref{fig:intro}. 
Hence, we propose an adversarial attack that systemically attacks the input image boundaries and increases the width of the adversarial boundaries until finding the Adversarial Example. 
We choose the input boundaries for two reasons: 
first, human vision tends to focus on the center of images and ignore the boundary, which is known as the center bias~\cite{center1,center2,center3}; 
second, although the input image boundaries usually do not overlap with the salient objects, they can be utilized to (i) improve the model performance as discussed in some special padding techniques of DNN models~\cite{padding1,padding2,padding3}; and (ii) encode the absolute position information in the semantic representation learning~\cite{position1,position2}.
Thus, image boundaries can have a desirable property in the imperceptible adversarial attack: image boundaries can be unimportant in human vision, but dominate a deep learning model.

Moreover , this study aims at studying the weakness of the DNN 
models from different perspectives such as correlating the boundaries to the Adversarial Examples regardless of what adversarial label the model produces. 
Henceforth, our experiments focus on the un-targeted attack and the white-box setup where the attacker has full access to the model. Our findings have the potential to help advance the understanding of Adversarial Examples and provide insights into what can be the reasons behind the existence of Adversarial Examples. Our contributions are three-manifold:

\begin{itemize}

\vspace{-0.5em}
\item Proposing a novel adversarial attack that systemically attacks the input image from the boundaries where the width of the attacked boundaries increases until the Adversarial Example is found. 
Our adversarial attack was effective with an average success rate of 95.2\% when attacking six CNN models and the ViT while only modifying less than 32\% of the input image content. 

\vspace{-0.5em}
\item	Attacking only the input image boundaries, so that improving the imperceptibility of the adversarial perturbations, which is supported by  
an average peak signal-to-noise ratio ($PSNR$) of $41.37$ in the experiments. 

\vspace{-0.5em}
\item Correlating the Adversarial Examples to the unimportant features (\textit{i.e.}, image boundaries) to provides a different perspective to understand the Adversarial Examples. 
We show the boundary width required to achieve a desired success rate, and how a model's attention changes when attacked by the proposed boundary adversarial examples.
\end{itemize}

\vspace{-0.25em}
The remainder of this paper is organized as follows. In Section~\ref{sec:literature}, we review and categorize the related work. Section~\ref{sec:implem} discusses our approach for our adversarial attack followed by evaluation results in Section~\ref{sec:results}. 
Finally, Section~\ref{sec:conclusion} concludes with the discussion on evaluation and highlights some of the future work in this sector.
\section{Related Work} \label{sec:literature}
This section briefly reviews the literature by dividing the adversarial perturbations into two main categories: Imperceptible Perturbations and Perceptible Perturbations.

\subsection{Imperceptible Perturbations}

At the early stage of the field, the Adversarial Examples~\cite{L-BFGS} were introduced to be imperceptible to human vision. 
Hence, the addition of an adversarial perturbation to the Clean Example has barely a human-vision effect on the Clean Example.
Based on the coverage of the Imperceptible Perturbations, there are two types of adversarial perturbations: Fully Coverage based and Partially Coverage based.

\vspace{-1.0em}
\subsubsection{Fully Coverage Based}
\vspace{-0.75em}
In this group, the adversarial perturbation is made to be of the same size as the Clean Example. 
Hence, each feature in the Clean Example will be affected by the corresponding feature in the adversarial perturbation. Szegedy \textit{et al.}~\cite{L-BFGS} introduced an expensive adversarial attack which was based on L-BFGS, and guaranteed to find the Adversarial Example. 
Then, Goodfellow \textit{et al.}~\cite{FGSM} proposed a cheap and effective adversarial attack called Fast Gradient Sign Method, FGSM. The sign function of the gradients in the FGSM helps determine the modification directions in the input's features. 
The FGSM used an $epsilon$ as quantity to equally change the features with their corresponding directions; however, the FGSM did not guarantee to produce the Adversarial Example and the attack depended on the value of $epsilon$. 
That is why Kurakin~\textit{et al.}~\cite{IFGSM} proposed an iterative FGSM where the input was kept being perturbed by a small $epsilon$ until the Adversarial Example was found. 
Moreover, Moosavi-Dezfooli~\textit{et al.}~\cite{deepfool} proposed an adversarial attack, DeepFool, which improved the imperceptibility of the adversarial perturbations by finding the smallest adversarial perturbation that reliably converted a Clean Example to its Adversarial Example. 


\vspace{-1.0em}
\subsubsection{Partially Coverage Based}
\vspace{-0.75em}
In this group, the adversarial perturbations change a subgroup of features in the input's space to find the Adversarial Example. 
For instance, Papernot~\textit{et al.}~\cite{jsma} proposed a Jacobian-based Saliency map Attack, JSMA, that could find the salient features in the input which if attacked, the Adversarial Example could be found. 
Different studies followed the JSMA to use the Jacobin-based saliency map in their adversarial attacks such as maximal Jacobian based Saliency map attack (MJSMT)~\cite{mjsmt}, and Probabilistic Jacobian based Saliency map attack (PJSMT)~\cite{pjsmt}. 
Also, Qian \textit{et al.}~\cite{visually} proposed an adversarial attack that used network explanations to find a small suitable semantic region in the input for adversarial modifications. 

Moreover, Wu~\textit{et al.}~\cite{boostingTransferability} proposed an adversarial attack that increased the transferability of the Adversarial Examples in the transfer-based black setting. 
They claim that Adversarial Examples crafted by a source model might not fool another model (target model) and that was due to the over-fitting to the source model. 
Hence, they proposed a procedure that used an attention mechanism to extract the important features to attack. 


\vspace{-0.5em}
\subsection{Perceptible Perturbations}
\vspace{-0.5em}
In this category, the perceptible perturbations occlude small areas in the Clean Example; hence, they are visually conspicuous to human vision but usually ignorable. 
Also, they come in different sizes and shapes. For example, Brown~\textit{et al.}~\cite{adversarailPatch} proposed an adversarial attack that could produce an adversarial patch when partially covering the Clean Example; it would fool the classifier. Karmon~\textit{et al.}~\cite{lavan} addressed the problem of the adversarial patch's size by introducing a small noise that can be localized in the Clean Example in a way it did not overlap with the salient object in that example, and still fool the DNN model. Evtimov\textit{et al.}~\cite{rp2} proposed Robust Physical World Attack, RP2, that could produce small adversarial stickers in the shape of road vandalism (such as camouflage art and graffiti). 
Hence, when such an adversarial sticker was physically attached to a road sign, the classifier would misclassify the sign. Sharif~\textit{et al.}~\cite{glassframe} introduced a glass frame that could fool the face identification model. 
Finally, Zajac~\textit{et al.}~\cite{edgeframe} proposed an adversarial frame that could be placed on the edges of a Clean Example and would fool both the image classifier and object detector. 
\vspace{-0.75em}
\section{The Proposed Boundary Attack}
\label{sec:implem}

\vspace{-0.5em}
Algorithm \ref{eddgeAttack} produces an Adversarial Example by attacking the input's boundaries while keeping the remaining part intact. 
We explain the algorithm by decomposing it into two loops: an outer loop and an inner loop. 
The former increases the boundary's width while the latter attacks the boundary through the iterative FGSM (I-FGSM).
Sections \ref{outerloop} and \ref{innerloop} detail the discussions loop respectively.

\begin{algorithm}[htp]
\caption{Boundary Attack}\label{eddgeAttack}
\begin{algorithmic}[1]
  
  \Input{$m$, $l_{m}$, $f_{\theta}$, $\epsilon$, $minimum$, where $m$ is the Clean Example assuming the format is a channel last, $l_{m}$ is the true label, $f_{\theta}$ is the model, $\epsilon$ is the initial epsilon, and $minimum$ is the lower bound for the epsilon.}
  \Output{$m'$ which is the Adversarial Example.}
  \State $w\leftarrow 1$ \hfill{/* initial width for edges .*/}  
  \While{$w<40$} \hfill{/* the outer loop.*/}
  \State $m'\leftarrow m$ \hfill{/* initial Adversarial Example.*/}
  \State $cnt\leftarrow 0$
   \While{$cnt<15$} \hfill{/* 15 is replaced with 50 for other attacks as explained in Section \ref{Benchmarks} .*/}
   \State $m'\leftarrow Boundary_{fgsm}(m', w, l_{m}, f_{\theta},  \epsilon) $ 
   \State $l_{m'} \leftarrow f_{\theta}(m')$
    \If{$l_{m}\not=l_{m'}$}
    \If {$PSNR > 40$ OR $\epsilon == minimum$}
    \State break outer loop.
  \EndIf
  \State break inner loop.
  \EndIf
  \State $cnt\leftarrow cnt +1 $
  \EndWhile \label{euclidendwhile1}
  \State $w\leftarrow w + 1$
  \State $\epsilon\leftarrow \epsilon* 0.75$
  
  \EndWhile\label{euclidendwhile}
  \State \textbf{return} $m'$
  
\end{algorithmic}
\end{algorithm}

\begin{figure}[t!]
\includegraphics[width=8.5cm]{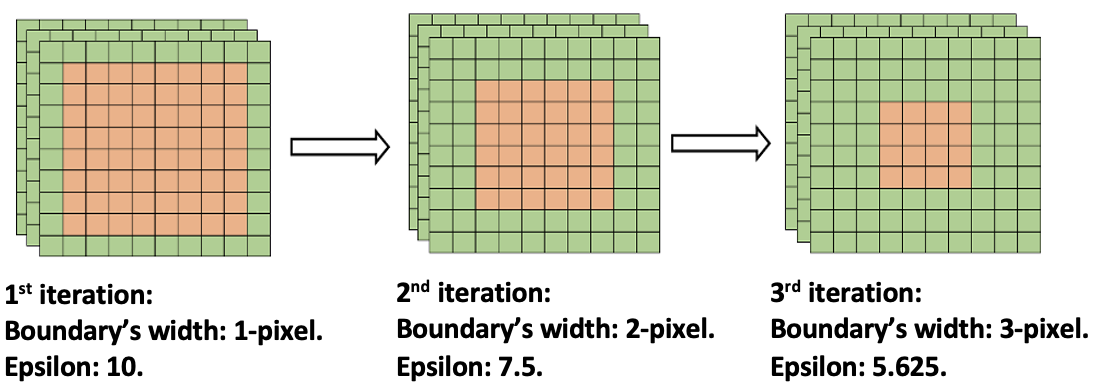} 
   \centering
   \caption{An example of input with a small shape of (8,8,3) to illustrate how the outer loop of Algorithm \ref{eddgeAttack} increases the width of the borders. For example, in the $1^{st}$ iteration, the width of the border is one-pixel (in green color), it becomes two-pixel in the $2^{nd}$ iteration attacks, and three-pixel in $3^{rd}$ iteration. Also, the epsilon decreases by 0.75 in each iteration. 
   }
   \vspace{-1.0em}
   \label{fig1}
\end{figure}

\vspace{-0.75em}
\subsection{The outer loop}
\label{outerloop}
\vspace{-0.5em}
The loop increases the boundary's width to be attacked in each iteration. It starts with one-pixel boundary (step $1$ in Algorithm~\ref{eddgeAttack}) and keeps the remaining area untouched. 
If the current boundary's width does not succeed in finding the Adversarial Example, the width will increase by one pixel in the next iteration (step $16$). 
Figure \ref{fig1} is an example to visually illustrate how the boundary's width increases in each iteration of the outer loop; the width is one-pixel in the $1^{st}$ iteration, two-pixel in the $2^{nd}$ iteration, and three-pixel in the $3^{rd}$ iteration. 
The outer loop keeps iterating until the inner loop signals a break for the outer loop (explained in Section \ref{innerloop}). 

For imperceptibility, the outer loop starts with an $epsilon$ value, $\epsilon$, that magnifies the signs of boundaries' gradients ( explained in Section \ref{innerloop}). 
Then, the $\epsilon$ will be decreased by a factor of $0.75$ in every iteration as shown in Figure \ref{fig1}; however, the decrement process will stop when reaching the lower bound for $\epsilon$ called $minimum$ in Algorithm \ref{eddgeAttack} (and not shown in Figure \ref{fig1}). 

\begin{algorithm}[htp]

\caption{$Boundary_{fgsm}$}\label{attack}
\begin{algorithmic}[1]
  
  \Input{$m'$, $w$, $l_{m}$, $f_{\theta}$, $\epsilon$, $\alpha$}
  \Output{$m''$}
  \State $g \leftarrow \nabla_{m'}f(\theta, m' , l_{m})$
  \State $g[ w:-w,  w:-w, : ] \leftarrow zeros$ 
  \State $p\leftarrow \epsilon * sign(g)$ 
  \State $m''\leftarrow m' + p$ 
  \State \textbf{return} $m''$
\end{algorithmic}
\end{algorithm}
\vspace{-1.0em}

\subsection{The Inner loop}\label{innerloop}

The inner loop is steps $5$ to $15$ in Algorithm~\ref{eddgeAttack}. 
It perturbs the $m'$ with the current $\epsilon$ value using $Boundary_{fgsm}$ function (as shown in step $6$) which is a version of the $FGSM$ attacking only the input's boundary. 
The details of the $Boundary_{fgsm}$ function is in Algorithm~\ref{attack} where step $1$ computes the model's gradients with respect to the input $m'$ given that the true label is $l_{m}$; step $2$ masks the middle part of the gradients $g$ by reassigning them $zeroes$ and keeps the boundary's gradients untouched, step $3$ is the adversarial perturbation where the sign is multiplied by the $\epsilon$ value. 
Lastly, step $4$ adds the adversarial perturbation to the $m'$ which gives an output that aims at maximizing the distance between the model's output class and the true label $l_{m}$.

The upper bound of the inner loop is set to be $15$ (step $5$ in Algorithm \ref{eddgeAttack}) which empirically keeps the perturbation imperceptible. 
The inner loop will keep perturbing the boundary using the $Boundary_{fgsm}$ function until it finds the Adversarial Example and it then breaks; otherwise, it reaches its upper bound with no success which in turn makes the outer loop go to the next iteration. 
Moreover, when the Adversarial Example is found, the inner loop will break the outer loop if either of the two following conditions is satisfied (as shown in step $9$ in Algorithm \ref{eddgeAttack}):

\begin{itemize}
\vspace{-0.8em}
\item When the ($PSNR$) is higher than a threshold. The threshold is set to be $40$ and chosen heuristically. 

\vspace{-0.8em}
\item When the $\epsilon$ reaches its lower bound $minimum$; that is because increasing the attack area while the epsilon value, $\epsilon$, is static, it never gives higher $PSNR$ than the smaller edges attacked by the same $\epsilon$.
\end{itemize}
\vspace{-0.8em}

These two conditions make the algorithm efficient at finding an Adversarial Example with the smallest perturbed area and the highest $PSNR$. 

\section{Experiments and Results} 
\label{sec:results}

This section discusses and analyzes our experimental setup, evaluation metrics, and quantitative results.

\subsection{Experimental Setup}

To test the proposed method, we attack the vision transformer (ViT)~\cite{vit}, and six widely applied CNN models including VGG16 and
VGG19~\cite{vgg}, ResNet50 and ResNet101~\cite{resnet}, and EfficientNetB1 and EfficientNetB2~\cite{fficientnet}. 
As for the CNN models, we train and evaluate the models using the Imagentte dataset~\cite{imagenette} which is a subset of the ImageNet dataset~\cite{imagenet} and has ten classes of tench, English springer, cassette player, chain saw, church, French horn, garbage truck, gas pump, golf ball, parachute. 
The Imagenette has a training dataset with $9,469$ images and a validation dataset with $3,925$ images; the size of the images is $(224,224,3)$ and they are also upsampled to $(240, 240, 3)$ and $(260, 260, 3)$ for the EfficientNetB1 and EfficientNetB2 respectively. 
We apply both the Imagenette and Tiny ImageNet~\cite{tiny} to the ViT; we use ViT(1) and ViT(2) to represent the ViT trained and tested on the Imagentte and Tiny ImageNet respectively. 
The Tiny ImageNet is also a subset of ImageNet, has $200$ classes, and has a training dataset with $100,000$ images and a validation dataset with $10,000$ images. The size of images in the Tiny ImageNet is $(64,64,3)$, and we upsample them to $(224,224,3)$.

We replace the last 1000-output layer with a
10-output layer in the CNN models and ViT(1) and to 200-output layer in the ViT(2). We use twenty epochs to fine-tune the output layer in the CNN models while fine-tuning the last transformer encoder and the output layer in the ViTs. 
The values of the input's pixels are in the range of [0,255] and [0,1] for the CNN models and the ViTs respectively. 
Hence, we heuristically choose the $\epsilon$ value and its minimums\ respectively to be $10$ and $3$ when attacking the CNN models while $0.02$ and $0.01$ when attacking the ViTs. 
Table~\ref{table:3} shows the accuracy of the models on the validation dataset. 
To ensure that the misclassifications are caused by the attack itself and not the model, we only attack the samples that are successfully classified from the validation dataset. 

\begin{table}[ht]
\centering
\begin{tabular}{|l|l|}
\hline
\textbf{Model} & \textbf{Accuracy} \\ \hline
VGG16 & 0.9156 \\ \hline
VGG19 & 0.9177 \\ \hline
ResNet50 & 0.957 \\ \hline
ResNet101 & 0.96 \\ \hline
EfficientNetB1 & 0.977 \\ \hline
EfficientNetB2 & 0.99 \\ \hline
ViT(1) & 0.986 \\ \hline
ViT(2) & 0.74 \\ \hline
\end{tabular}
\vspace{0.25em}
\caption{The accuracy of six CNN models on the validation  datatset of Imagenette. The accuracy of ViT(1) and ViT(2) are respectively on the validation  datatset of Imagenette and the validation  datatset of Tiny ImageNet.}
\label{table:3}
\end{table}
\vspace{-1.0em}


\subsection{The Evaluation Metrics}\label{metrics}
Four evaluation metrics are used in the experiments: Success Rate ($SR$), Mean Squared Error ($MSE$), Mean Absolute Error ($MAE$), and the $PSNR$. The $SR$ can be defined as followers:

\begin{equation} \label{sr}
SR(f_{\theta},M)= \frac {1}{T}\sum_{m\in M}1_{\bigl\{ f_{\theta}(m')\neq l_{m} \bigr\}},
\end{equation}

\noindent where $f$ is a model parameterized by $\theta$, $M$ represents all the images in the validation dataset that are correctly classified by the $f$, $T$ is the number of images in $M$, $l_m$ is the true label for the Clean Example $m$, and $m'$ is the perturbed Example crafted from the $m$. 

Moreover, the computations of the $MSE$, $MAE$, and $PSNR$ only consider $N\in M$ where $N$ represents images in the $M$ that are both correctly classified by $f_\theta$ and successfully attacked by our adversarial attack. The $MSE$, $MAE$, and $PSNR$ can be respectively defined by the following equations:

\begin{equation} \label{mse}
MSE(N)= \frac{1}{T}\sum_{n\in N}{mean((n - n')^2)},
\end{equation}

\begin{equation} \label{mae}
MAE(N)= \frac{1}{T}\sum_{n\in N}{mean(|n - n'|)},
\end{equation}

\begin{equation} \label{psrn}
PSRN(N)= \frac{1}{T}\sum_{n\in N} 20* \log{(\frac{max}{\sqrt{mean((n - n')^2)}})}, 
\end{equation}

\noindent where $T$ is the number of images in $N$, $n$ and $n'$ are the Clean Example and its Adversarial Example, $max$ is the highest possible value in the $N$, and the $mean$ function is the reduced mean. 
The purpose of using $MSE$ and $MAE$ is to show the difference between the Clean Example and its Adversarial Example from two different evaluation perspectives. Besides the $MSE$ and $MAE$, the $PSNR$ is used to show the ratio of the possible highest power of a signal to the power of that noise that corrupts that signal.

%

Finally, we investigate the model’s attention using the Grad-CAM~\cite{gradcam} for the Clean Example and its Adversarial Example, and how the adversarial perturbations can change the model's attention significantly. 
Hence, this experiment can correlate the unimportant features (\textit{i.e.}, input image boundaries)
with the Adversarial Examples; this correlation hopefully helps advance our understanding of the reasons behind the existence of the Adversarial Examples. 

\begin{figure}[t!]
  \centering
  \includegraphics[width=0.7\linewidth]{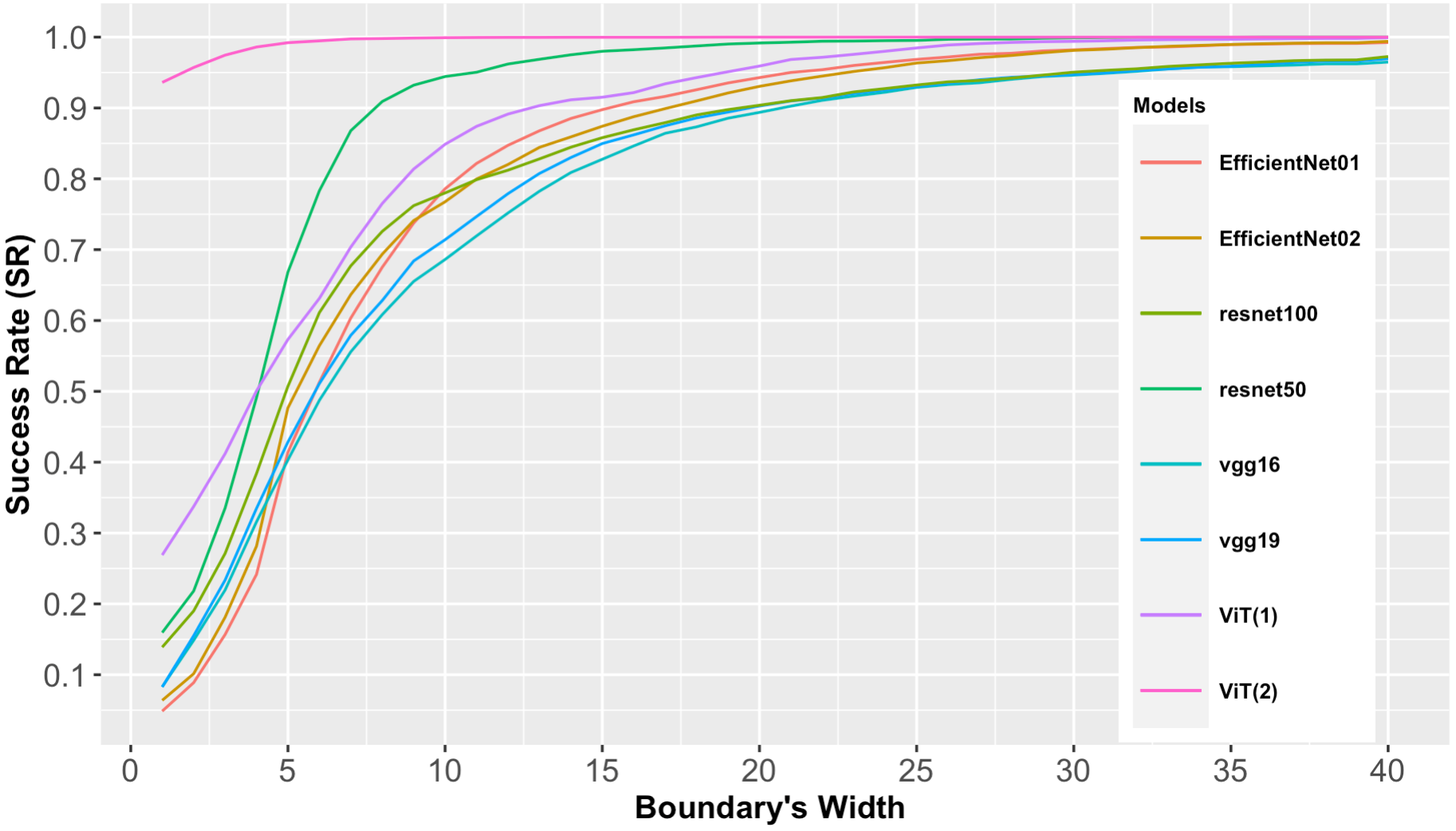}
  \caption{Our attack's $SR$s when attacking six different models and two ViTs where ViT(1) trained on Imagenette while ViT(2) is the same transformer but trained on the Tiny ImageNet.}
  \vspace{-1.5em}
  \label{fig:SR}
\end{figure}

\subsection{The Quantitative Results}
\subsubsection{Correlating the $SR$ with the boundary's width}
Figure~\ref{fig:SR} shows the comparison of our adversarial attack against the six CNN models and the two ViTs in terms of the $SR$ and the boundary's width. 
The ViT(2) is the most vulnerable model where the attack starts with a high $SR$ and achieves around 99.0\% with less than a 5-pixel boundary attack in all the $M$ dataset's images. 
This highest $SR$ is attributed to the large number of classes in the Tiny ImageNet which makes it easier for our adversarial attack to confuse the ViT model. 
Then, the ResNet50 model shows the second highest vulnerability; the ViT(1) becomes comparable to the Resnet50 when attacked with a 25-pixel boundary and both achieve an average $SR$ of $99.9\%$ at the 40-pixel boundary. 
Also, the attack's $SR$s against EfficientNetB1 and EfficientNetB2 converge with the other highest $SR$s at the boundaries of 30-pixel width. 
Lastly, the ResNet101, VGG19, and VGG16 are comparable to each other at the 20-pixel boundary, and they also achieve an average $SR$ of $96.9\%$ at the 40-pixel boundary.

\begin{figure}[t!]
  \centering
  \includegraphics[width=0.7\linewidth]{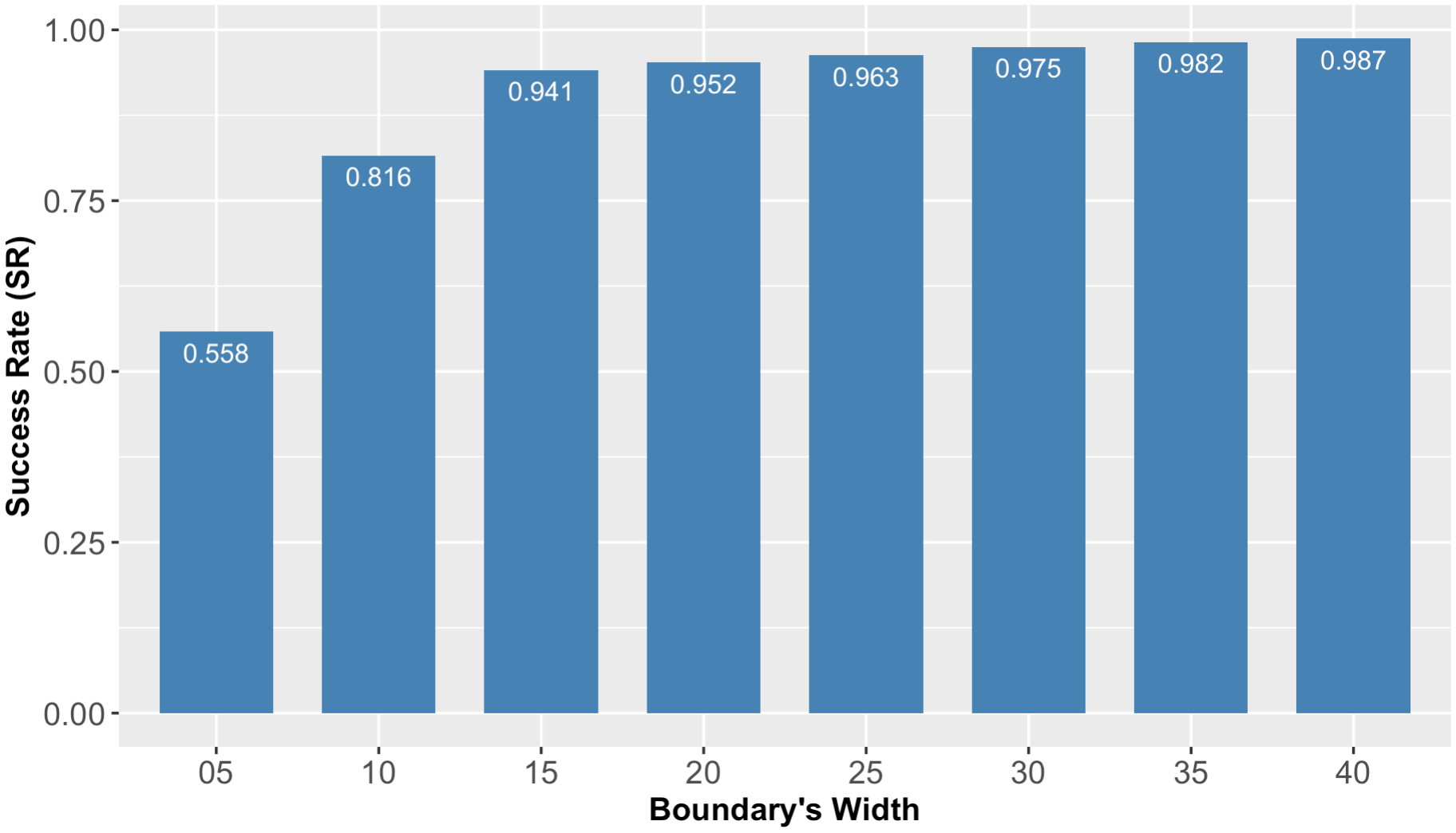}
  \caption{The relationship between the boundary's width and the Success Rate ($SR$). The $SR$ in each bar is an average of the $SR$s over all the six models and the two ViTs.}
  \vspace{-1.5em}
  \label{fig:SRvsW}
\end{figure}

Moreover, Figure \ref{fig:SRvsW} shows the relation between different widths for the adversarial boundaries and the $SR$. 
The purpose of this analysis is to show how much possible the attacker can produce an Adversarial Example in terms of the perturbation amount. 
For example, the attacker only needs a 5-pixel boundary ($8\%$ of the image content) to achieve a success rate of $55.8\%$, a 10-pixel boundary ($17\%$ of the image content) to achieve an average $SR$ of $81.6\%$, a 20-pixel boundary ($32\%$ of the image content) to achieve an average $SR$ of $95.2\%$, a 30-pixel boundary ($46\%$ of the image content) to achieve an average $SR$ of $97.5\%$, and a 40-pixel boundary ($58\%$ of the image content) to achieve an average $SR$ of $98.7\%$. 
It is evident that as we increase the width of the boundary, the success rate increases. Hence, the attacker can increase the width of the boundary to increase the possibility of getting an Adversarial Example.

\vspace{-1.0em}
\subsubsection{Comparative Study}
\label{Benchmarks}
\vspace{-0.5em}

Since our adversarial attack integrates two concepts of attacking unimportant features (\textit{e.g.}, boundaries) and imperceptibility, it is hard to directly compare our work with the related work in Section~\ref{sec:literature}. 
However, we modify Algorithm~\ref{eddgeAttack} to re-implement three different attacks: the adversarial patch of~\cite{lavan}, the adversarial frame of~\cite{edgeframe}, and attacking the whole example by I-FGSM~\cite{IFGSM} (henceforth, we call it $Whole$ attack). Then, we compare our attacks with these three attacks using the evaluation metrics mentioned in Subsection~\ref{metrics}.

\begin{table*}[!h]
\centering
\begin{tabular}{|l|l|l|l|l|l|l|l|}
\hline
\textbf{Attacks} & \textbf{Metrics\textbackslash{}Models} & \textbf{VGG16} & \textbf{VGG19} & \textbf{ResNet50} & \textbf{ResNet101} & \textbf{EfficientNetB1} & \textbf{EfficientNetB2} \\ \hline
\multirow{4}{*}{\textbf{Patch}} & \textit{SR} & 0.88 & 0.88 & 0.88 & 0.64 & 0.85 & 0.75 \\ \cline{2-8} 
 & \textit{MAE} & 4.5 & 4.4 & 4.4 & 4.5 & 3.9 & 3.33 \\ \cline{2-8} 
 & \textit{MSE} & 595.4 & 593.6 & 591.8 & 591.8 & 517.7 & 517.8 \\ \cline{2-8} 
 & \textit{PSNR} & 38.86 & 38.86 & 39.12 & 40.16 & 36.28 & 35.61 \\ \hline
\multirow{4}{*}{\textbf{Frame}} & \textit{SR} & 0.96 & 0.96 & 0.997 & 0.95 & 0.85 & 0.96 \\ \cline{2-8} 
 & \textit{MAE} & 7.8 & 7.7 & 7.7 & 7.6 & 7.2 & 6.6 \\ \cline{2-8} 
 & \textit{MSE} & 1039.9  & 1042.8  & 1038.9  & 1029.3 & 965.7 & 897.8 \\ \cline{2-8} 
 & \textit{PSNR} & \textbf{40.01} & \textbf{40.6} & 38.34 & 37.2 & 32.36 & 33.07 \\ \hline
\multirow{4}{*}{\textbf{Whole}} & \textit{SR} & \textbf{0.995} & \textbf{0.9997} & \textbf{1.0} & \textbf{0.9997} & \textbf{1.0} & \textbf{1.0} \\ \cline{2-8} 
 & \textit{MAE} & 3.4  & 3.5 & 2.9  & 3.4 & 3.5  & 3.4 \\ \cline{2-8} 
 & \textit{MSE} & 19.4 & 19.5 & 9.9 & 18.2 & 20.0  & 18.8 \\ \cline{2-8} 
 & \textit{PSNR} & 36.26 & 36.31 & 38.31 & 36.31 & 35.69 & 36.00 \\ \hline
\multirow{4}{*}{\textbf{Ours}} & \textit{SR} & 0.96 & 0.97 & \textbf{1.0} & 0.97 & 0.99 & 0.99 \\ \cline{2-8} 
 & \textit{MAE} & \textbf{0.85} & \textbf{0.83} & \textbf{0.52} & \textbf{0.68} & \textbf{0.99} & \textbf{0.93} \\ \cline{2-8} 
 & \textit{MSE} & \textbf{8.14} & \textbf{8.02} & \textbf{5.42} & \textbf{6.61 } & \textbf{12.39} & \textbf{12.4} \\ \cline{2-8} 
 & \textit{PSNR} & 39.79 & 39.87 & \textbf{41.0} & \textbf{40.42} & \textbf{38.14} & \textbf{38.36} \\ \hline
\end{tabular}
\vspace{0.25em}
\caption{Six models (first row) are attacked by four different attacks (first column). Four metrics (second column) including $SR$, $MAE$, $MSE$, and $PSNR$ are used to compare the attacks.}
\label{table:1}
\end{table*}

\begin{table}[!h]
\centering
\begin{tabular}{|l|l|l|l|l|l|}
\hline
\textbf{Models} & \textbf{Attacks} & \textbf{SR} & \textbf{MAE} & \textbf{MSE} & \textbf{PSNR} \\ \hline
\multirow{4}{*}{\textbf{ViT (1)}} & Patch & 0.90 & 0.12 & 0.03 & 20.3 \\ \cline{2-6} 
 & Frame & 0.99 & 0.09 & 0.02 & 23.5 \\ \cline{2-6} 
 & Whole & \textbf{1.0} & 0.13 & 0.27e-3 & 36.4 \\ \cline{2-6} 
 & Ours & 0.999 & \textbf{0.003} & \textbf{0.13e-3} & \textbf{39.5} \\ \hline
\multirow{4}{*}{\textbf{ViT (2)}} & Patch & 0.99 & 0.05 & 0.013 & 31.6 \\ \cline{2-6} 
 & Frame & 0.999 & 0.043 & 0.013 & 35.0 \\ \cline{2-6} 
 & Whole & \textbf{1.0} & 0.01 & 0.13e-3 & 38.95 \\ \cline{2-6} 
 & Ours & \textbf{1.0} & \textbf{0.001} & \textbf{3.1e-5} & \textbf{46.8} \\ \hline
\end{tabular}
\vspace{0.25em}
\caption{ViT (1) is trained on the Imagenette while ViT (2) is trained on the Tiny ImageNet. The evaluation metrics ($SR$, $MAE$, $MSE$, and $PSNR$) are the same ones in Table~\ref{table:1}.}
\vspace{-1.5em}
\label{table:2}
\end{table}


The modifications are as follows: First, at the beginning of Algorithm \ref{eddgeAttack}, we set the patch's size to be $(50,50, 3)$ and place it at the top left corner of the Clean Example $m$; the frame's width is set to be 5-pixel and occludes the original boundaries; the patch and frame are initialized randomly. 
Secondly, the outer loop is removed, and only the inner loop is included in Algorithm \ref{eddgeAttack}. 
The upper bound for the inner loop is set to be 50 (more than three times of our attack's upper bound); 
the $\epsilon$ value is set to be static (\textit{i.e.}, equal to the lower bound used in our attack; for example $3$ and $0.01$ for the CNN models and ViTs respectively).  
Lastly, step $2$ in Algorithm \ref{attack} masks the gradients that are out of the patch and frame areas, and no mask in the $Whole$ attack since I-FGSM attacks the whole image. 

Table~\ref{table:1} shows the comparison of our attack with the three attacks (adversarial patch, adversarial frame, and the $Whole$ attack) when attacking the six CNN models. 
Also, Table~\ref{table:2} shows the same comparison when attacking the ViTs. 
In both tables, we can observe that the perturbations (differences between the Clean Examples and Adversarial Examples in terms of the $MSE$ and $MAE$) in our attack are significantly lower than all the other attacks against the six models and the ViTs. Moreover, the $PSNR$s are shown to be higher for our attack compared to the other attacks against the CNN models and ViTs unless the VGGs attacked by the Frame are higher than ours; however, the average $PSNR$ for our attack is $41.37$ which is higher than all the other attacks. Also, The $SR$s show that our attacks can be comparable to the $Whole$ attack and much higher than the other two attacks (adversarial patch and frame). 
Remarkably, we set the upper bound of the inner loop in Algorithm~\ref{eddgeAttack} for the three attacks ($Whole$ attack, adversarial patch, and frame) to be three times more than our attack for fair comparison (\textit{i.e.}, $15$ to $50$), otherwise the $SR$ for the three attacks would increase if we would increase their upper bounds above $50$.

Finally, if we compare the $SR$s for all the attacks in Table~\ref{table:1} and Table~\ref{table:2} against the six CNN models and the ViTs respectively, we can observe that the ViTs are more vulnerable than the CNN models for all the attacks.

\vspace{-0.5em}
\subsection{Spatial Attention Analysis}
\vspace{-0.25em}

\begin{figure}[t!]
\centering
\includegraphics[width=0.75\linewidth]{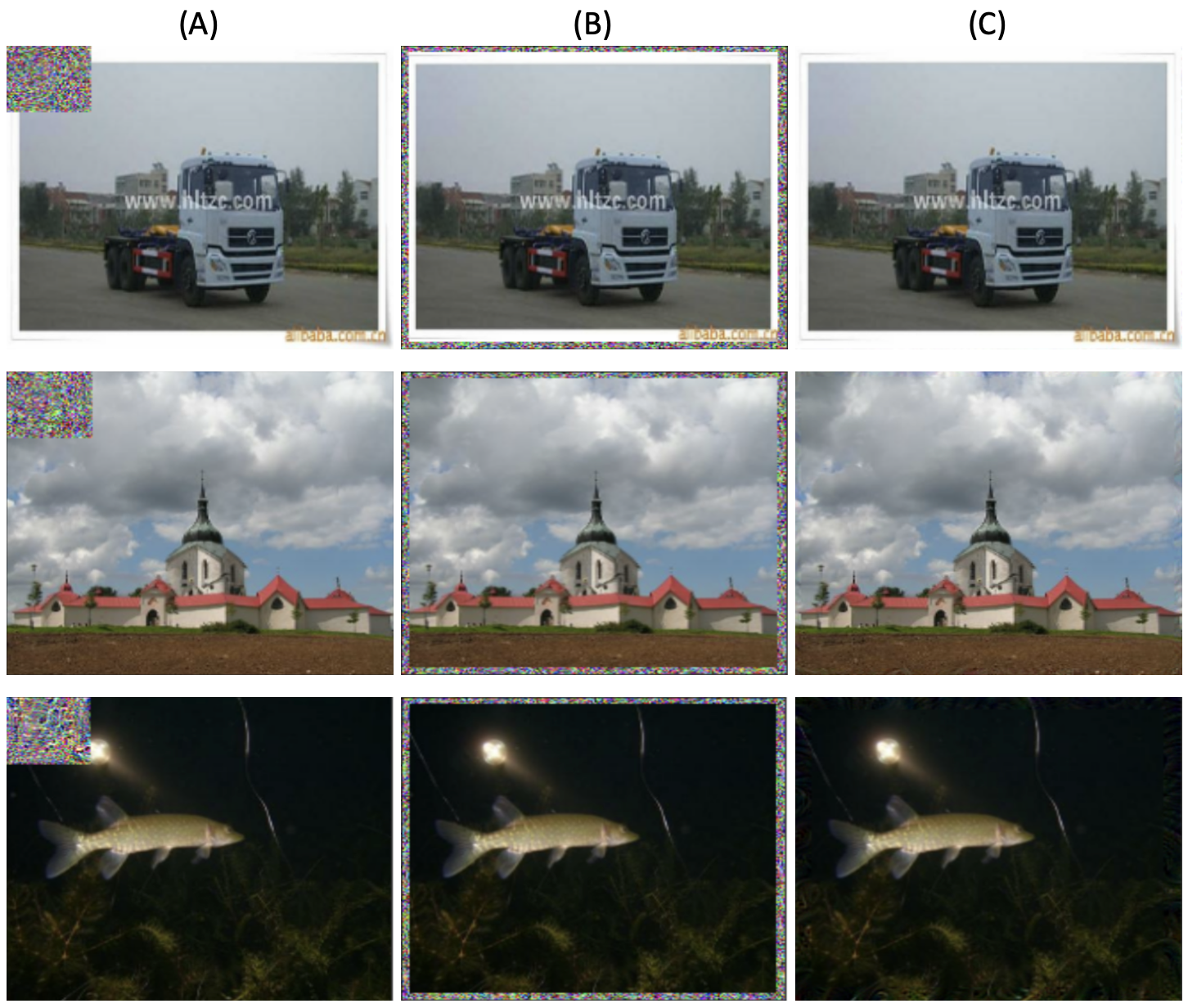} 
   \caption{Three samples of successful attacks. The $1^{st}$ row is a sample that can be attacked by a one-pixel boundary attack as shown in column (C); while attacking the same sample by the patch and frame attacks are respectively shown in columns (A) and (B) in the same row. Similarly, the $2^{nd}$ row and $3^{rd}$ row represent the five and ten boundary attacks respectively. The labels are explained in the model's attentions in Figure~\ref{fig:gradcam}.}
   \vspace{-2.0em}
   \label{fig:samples}
\end{figure}

\begin{figure*}[t]
   \centering
   \includegraphics[width=520pt,height=250pt]{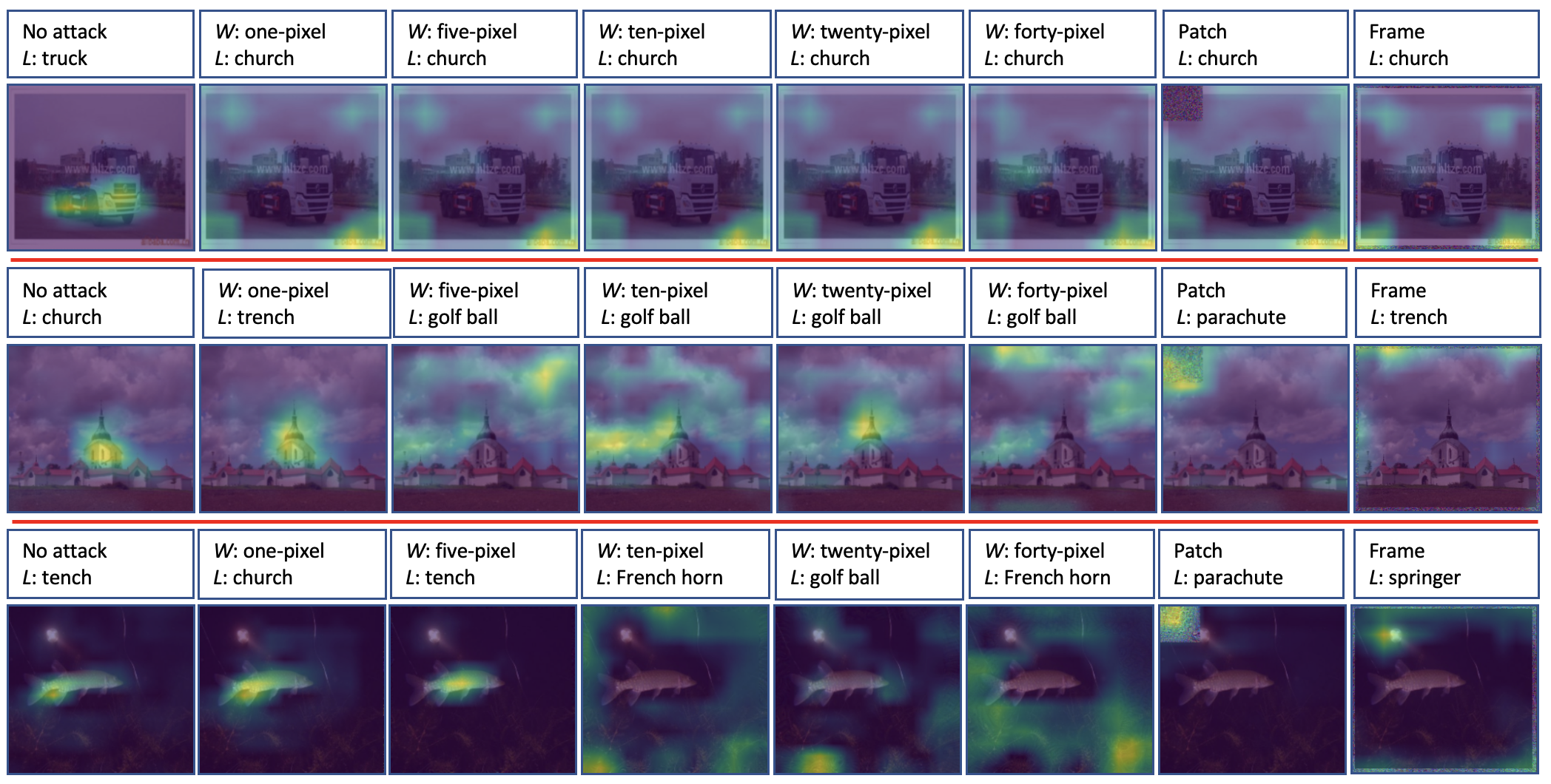}
   \caption{The model's attentions for the three samples in Figure~\ref{fig:samples}. The $1^{st}$ row is the sample that can be successfully attacked with only one-pixel boundary attack, the $2^{nd}$ row is the sample that can be successfully attacked by five-pixel boundary attack, and the $3^{rd}$ row  is the sample that can be successfully attacked by ten-pixel boundary attack. The $1^{st}$ column in each row is the model’s attention for the Clean Example, the $2^{nd}$ column to the $6^{th}$ column are one-pixel, five-pixel, ten-pixel, twenty pixel, and forty-pixel boundary attacks for the same sample in that specific row, the $7^{th}$ and $8^{th}$ are the model's attentions for the patch and frame attacks respectively. The $W$ means the boundary's width and $L$ means the assigned label by the model.}
   \vspace{-1.5em}
   \label{fig:gradcam}
\end{figure*}

In this experiment, we analyze how the adversarial perturbations at the boundaries change the model’s attention of where to look for the features that mainly determine the model’s output. 
We sample three successful attacks from the $N$ dataset representing three different boundary's widths (\textit{i.e.}, one-pixel boundary attack as in the $1^{st}$ row in Figure \ref{fig:gradcam}, five-pixel boundary attack as in the $2^{nd}$ row, and ten-pixel boundary attack as in the $3^{rd}$ row). 
For each sample, we show the model's attention for the Clean Example as in the $1^{st}$ column in Figure \ref{fig:gradcam}, while the $2^{nd}$ column to the $6^{th}$ columns show the model's attentions for one-pixel, five-pixel, ten-pixel, twenty-pixel, and forty-pixel boundary attacks respectively for the same sample, and the $7^{th}$ and $8^{th}$ are the attentions for the patch and frame attack respectively. Moreover, we show the three attacks (patch, frame, and our attack) for the three samples in Figure~\ref{fig:samples}

The goal of showing our attack with various boundary widths for each sample is to investigate if the model's attention for a successful attack will significantly change if the boundary's width increases. 
For example, the second row in Figure~\ref{fig:gradcam} represents a successful five-pixel boundary attack; hence, the model's attention to the unsuccessful attack in the $2^{nd}$ column (one-pixel boundary attack) still signifies features from the main object in the example (church) similar to the model's attention for no attack in the $1^{st}$ column with a small change in the size and shape of the attention. 
However, the five-pixel boundary attack for the same sample as in the $3^{rd}$ column is successful and made the model's attention totally different, and only small features from the main object are included in the attention; also, the attention changes as the boundary's width increases; but the model still gives the same adversarial label for the different boundary's widths attack against this specific sample. 

Similarly, for the sample in the $3^{rd}$ row in Figure \ref{fig:gradcam}), the model's attention does not significantly change when our attacks (i.e., one-pixel and five-pixel boundary) are not successful; 
however, our attack against this sample becomes successful when using a ten-pixel boundary (as in the $4^{th}$ column) and the attention significantly changes. 
Also, increasing the width of the boundary in this sample after a successful attack, changes the model's attention and the model give different adversarial labels (i.e., $5^{th}$ column and $6^{th}$ column in the last row in \ref{fig:gradcam}).



Lastly, we show the model's attentions for these three samples when attacked by the adversarial patch and frame as in the $7^{th}$ and $8^{th}$ columns in Figure \ref{fig:gradcam}. 
It is evident that in the second and third samples, the patches and frames become the model's attention; 
the adversarial labels are different from our attack. 
However, the patch and frame used to attack the first sample are also successful, the model gives the same adversarial label as our attack, and they changes the model's attention but the patch and frame themselves are not the model's attention. 

Interestingly, we observe that If the sample does not need a strong perturbation as in the case of the $1^{st}$ sample, all the attacks agree on the model's attention and the adversarial label. 
But, if the sample needs a strong adversarial perturbation as in the $2^{nd}$ and $3^{rd}$ samples, the patch and frame become the attention of the model and do not change the image context; while our boundary attacks can turn on some areas that are not in the boundary and make them significant. 
We conjecture that our attack perturbs the image context and changes the relationships/contrast between different parts of the image so that indirectly forces the model's attention to certain areas. We will further investigate this finding in future work.    
\section{Conclusion and Future Research Directions} 
\label{sec:conclusion}

This paper proposes an imperceptible adversarial attack from the input image boundaries. 
The proposed adversarial attack is shown to be effective when attacking six different CNN models and the Vision Transformer (ViT) with an average success rate of 95.2\% when modifying only 32\% of the content, mainly from the input boundaries which usually are ignored by the human vision and do not overlap with the salient objects in the input.

Our experimental results of attacking from the image boundaries align with related discoveries in the literature that the performance of CNN models can be dominated by the boundaries.
We show that the cutting-edge ViT models are also vulnerable to our proposed boundary adversarial attacks. 
We find that the ViT model is sensitive to the adversarial perturbations; it is more vulnerable to our attack than the CNN models. 
We provide some correlation analysis, such as showing how much boundary's width is required to achieve a desired success rate and how the model's attention changes when attacked by our adversarial attack with different boundary's widths. 
Our findings can potentially advance the understanding of Adversarial Examples and provide a different perspective on how Adversarial Examples can be constructed.

Lastly, one research direction to further this study is to determine which boundary in the input image is more dominating in the attacks; we suspect the need for equal adversarial perturbations in all the input boundaries. 
One possible solution for such an attack is to combine the saliency maps or Jacobian values of the boundaries in attacks.


\bibliographystyle{unsrt}  
\bibliography{egbib}

\end{document}